\documentclass[10pt,twocolumn,letterpaper]{article}

\usepackage[pagenumbers]{cvpr} % To force page numbers, e.g. for an arXiv version

\definecolor{cvprblue}{rgb}{0.21,0.49,0.74}
\usepackage[pagebackref,breaklinks,colorlinks,allcolors=cvprblue]{hyperref}
\usepackage{algorithm,algorithmic}
\usepackage{multirow}
\usepackage{amsfonts}
\usepackage{bbding}
\usepackage{amsmath}
\usepackage[table]{xcolor}
\usepackage{color}
\usepackage{amssymb}
\usepackage{bbm}
\usepackage{bm}
\usepackage[accsupp]{axessibility}

\def\paperID{85} % *** Enter the Paper ID here
\def\confName{CVPR}
\def\confYear{2026}

\title{Boosting Robust AIGI Detection with Lora-based Pairwise Training}

\author{
Ruiyang Xia$^{1,2}$ \quad Qi Zhang$^1$ \quad Yaowen Xu$^1$ \quad Zhaofan Zou$^1$ \quad Hao Sun$^1$\thanks{Corresponding author.} \\ Zhongjiang He$^1$ \quad Xuelong Li$^1$\protect\footnotemark[1] \\
$^1$Institute of Artificial Intelligence (TeleAI), China Telecom \quad $^2$Xidian Univerisity \\
\tt\small ryon@stu.xidian.edu.cn \\ \tt\small \{zhangq139, xuyw1, zouzhf41, sunh10, hezj\}@chinatelecom.cn  \\  \tt\small xuelong\_li@ieee.org
}

\begin{document}
\maketitle
\begin{abstract}
The proliferation of highly realistic AI-Generated Image (AIGI) has necessitated the development of practical detection methods. While current AIGI detectors perform admirably on clean datasets, their detection performance frequently decreases when deployed "in the wild," where images are subjected to unpredictable, complex distortions. To resolve the critical vulnerability, we propose a novel LoRA-based Pairwise Training (LPT) strategy designed specifically to achieve robust detection for AIGI under severe distortions. The core of our strategy involves the targeted fine-tuning of a visual foundation model, the deliberate simulation of data distribution during the training phase, and a unique pairwise training process. Specifically, we introduce distortion and size simulations to better fit the distribution from the validation and test sets. Based on the strong visual representation capability of the visual foundation model, we fine-tune the model to achieve AIGI detection. The pairwise training is utilized to improve the detection via decoupling the generalization and robustness optimization. Experiments show that our approach secured the 3th placement in the NTIRE Robust AI-Generated Image Detection in the Wild challenge.
\end{abstract}    
\section{Introduction}
\label{sec:introduction}

\par The rapid and unprecedented evolution of generative artificial intelligence over the past decade has fundamentally transformed the landscape of digital media creation. With the advent of highly sophisticated architectures—most notably Diffusion Models \cite{ddim,DiT} and advanced Generative Adversarial Networks (GANs) \cite{style}, the synthesis of high-fidelity visual content has become remarkably accessible. These models are capable of generating images with a level of photorealism that seamlessly bridges the uncanny valley, rendering them virtually indistinguishable from pristine, authentically captured photographs to the naked human eye. While this technological development has democratized artistic expression and accelerated workflows in creative industries, it has simultaneously introduced a spectrum of severe societal vulnerabilities. The malicious deployment of AI-Generated Image (AIGI) to propagate disinformation, fabricate non-consensual deepfakes, and bypass digital authentication systems poses a critical threat to information integrity. Consequently, the development of highly accurate, generalized, and robust AIGI detection frameworks has emerged as a paramount objective within the computer vision community.

\begin{figure}
\centering
\includegraphics[width=\linewidth]{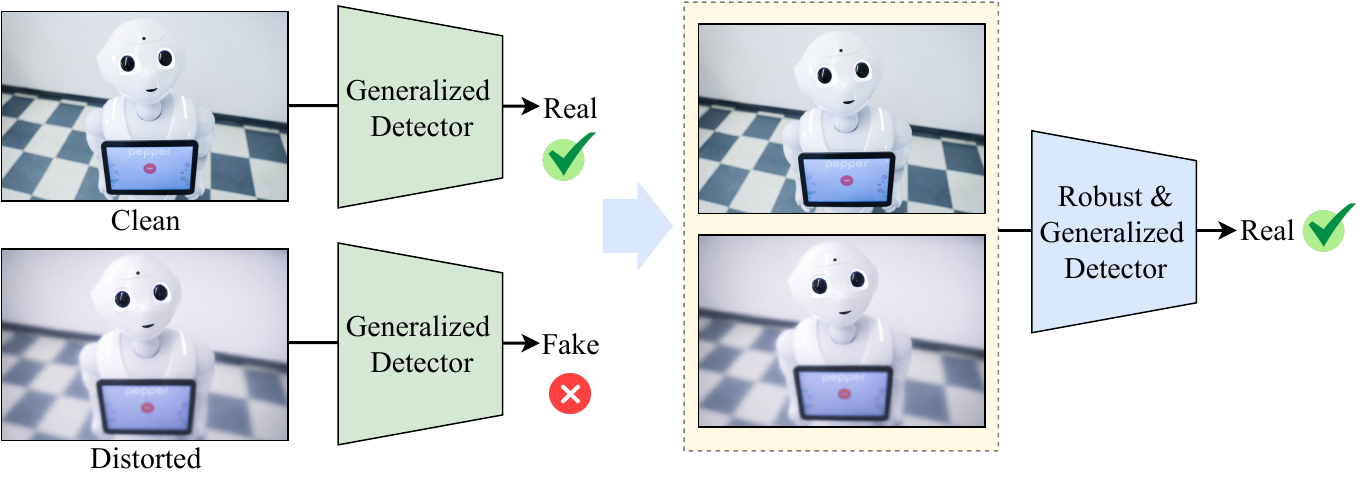}
\caption{Previous AIGI detection methods mainly consider the generalization performance under different generation models. However, when the images are distorted due to the transmission, the detectors will make false judgments. To make the detector practical in the downstream scenario, it is urgent to consider a robust and generalized detector.}
\label{f_1}
\end{figure}

\par Despite significant strides in AIGI detection \cite{wang2020cnn,oddn,shi2025shield}, a formidable challenge remains: the pervasive domain gap between the controlled environments in which detectors are trained and the chaotic, unconstrained realities of the real-world environments. As illustrated in Fig.~\ref{f_1}, the vast majority of contemporary AIGI detection methods are trained and evaluated on clean, uncompressed datasets directly outputted by generative models. However, when these generated images are disseminated across social media platforms, they are subjected to many complex and multi-stage degradations. These "in-the-wild" distortions—which routinely include aggressive JPEG compression, arbitrary scaling, random cropping, blurring, and the introduction of various noise—act as a severe interference to the detectors. These distortions impact the high-frequency checkerboard artifacts, microscopic pixel-level anomalies, and subtle blending inconsistencies that traditional detectors rely upon for classification. As a result, a model that achieves near-perfect accuracy on a pristine test set will frequently experience a catastrophic drop in performance when confronted with real-world, degraded imagery.

\par To systematically address this critical vulnerability, we introduce the LoRA-based Pairwise Training (LPT) strategy specifically engineered to achieve a robust and generalized detector for AI-Generated images under severe distortions. Developed within the context of the NTIRE 2026 Robust AI-Generated Image Detection in the Wild challenge, our framework abandons the reliance on fragile high-frequency artifacts in favor of extracting resilient, high-level semantic representations. The proposed LPT strategy achieves this through a synergistic three-step process: the parameter-efficient fine-tuning of a large-scale visual foundation model, the deliberate and aggressive simulation of data distributions during the training stage, and a novel pairwise training strategy.

\par Firstly, we resolve the data distribution shift via identifying that standard baseline distortions are insufficient to mimic real-world scenarios. By analyzing the validation and public test set provided by the organizers, we additionally introduce different image distortions with different combinations and amplify these degradations by setting the mean of the applied Gaussian sampling distribution to 3. Besides, we further incorporate a random crop and resize algorithm to better fit the size diversity. Secondly, we utilize the formidable zero-shot capabilities of the visual foundation model as our detector backbone. Instead of fully fine-tuning this massive architecture—which risks catastrophic forgetting of its generalized representations—we thus employ Low-Rank Adaptation (LoRA) within the multi-head self-attention (MHSA) and feed-forward network (FFN) blocks of each visual transformer block. Finally, to ensure that our model does not sacrifice its baseline accuracy on pristine images while learning to navigate degradations, we implement a pairwise optimization strategy. By explicitly pairing clean and distorted samples within the same training batch and correcting the extracted features through a dedicated feed-forward network, we force the model to map corrupted inputs back to their pristine semantic space. Consequently, our proposed approach ultimately secured the third placement in the NTIRE 2026 challenge\cite{ntire26aigendet}.

\par The contributions of this paper are summarized as:

\begin{itemize}
\item We utilize a large visual foundation model as our detector backbone, employing parameter-efficient fine-tuning to boost detection accuracy without compromising the model's inherent generalization capabilities.

\item We introduce distortion and size simulation during the training stage to better fit the data distribution from the validation and test sets.

\item We incorporate a novel training strategy that pairs clean and distorted samples within each training batch, guided by a joint loss function to enforce feature consistency.

\item Experiments demonstrate the effectiveness of each proposed component. Consequently, LPT secured the third placement in the NTIRE Robust AI-Generated Image Detection in the Wild challenge.

\end{itemize}
\section{Related Work}
\label{sec:related_work}
\subsection{Detection Paradigms for AI-Generated Image}

Early digital forensic techniques primarily focused on spatial-domain analysis, relying on Convolutional Neural Networks (CNNs) to detect local inconsistencies, anomalous color channel distributions, and checkerboard artifacts introduced by standard upsampling operations \cite{wang2020cnn,marra2019gans,yu2019attributing,LiuIJCAI,cheng2025fair}. Subsequent research pivoted toward the frequency domain, utilizing Discrete Cosine Transforms (DCT) or Fourier analysis to identify unnatural spectral peaks that are rarely present in natural photography \cite{durall2020watch,frank2020leveraging,zhang2019detecting}. While these methods proved highly effective in controlled environments, they exhibited severe fragility when exposed to standard image processing operations. For instance, a simple JPEG compression pass acts as a low-pass filter, effectively erasing the high-frequency clues that frequency-based detectors depend upon \cite{dong2022think,ojha2023towards}. Recognizing this limitation, recent state-of-the-art approaches have begun treating AIGI detection as a global semantic anomaly detection task, shifting the focus from microscopic pixel analysis to macroscopic contextual understanding \cite{ojha2023towards,zheng2024breaking,tan2025anomreason}.

\subsection{Visual Foundation Models for Digital Forensics}

The introduction of Vision-Language foundation models, most notably Contrastive Language-Image Pre-training (CLIP), has provided a powerful model for computer vision tasks \cite{clip,jia2021align,zhai2022lit,singh2022flava}. Pre-trained on massive datasets of hundreds of millions of image-text pairs, foundation models learn highly robust, generalized feature representations that align visual concepts with semantic language \cite{clip,jia2021align}. By scaling up these architectures and employing advanced training methodologies, modern visual foundation models achieve superior feature extraction capabilities and enhanced zero-shot transferability. Because these massive models have already learned a vast approximation of the natural visual world during their pre-training phase, they serve as an ideal baseline for detecting the subtle semantic anomalies and physical impossibilities often present in AIGI \cite{ojha2023towards,cui2025forensics,sun2025general}.

\begin{figure*}[t]
\centering
\includegraphics[width=\linewidth]{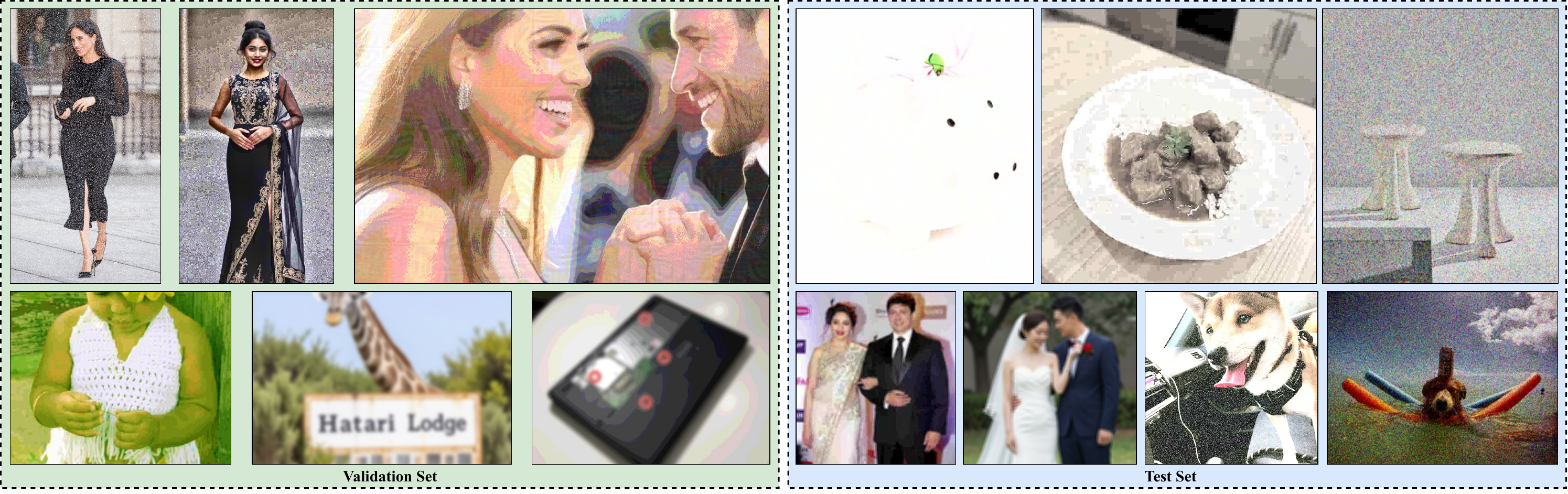}
\caption{\textbf{Distorted Images Illustration.} Irrespective of whether the images are real or fake, all images are subjected to multiple distortions with high-level strength, such as Gaussian blur, quantization, impulse noise, brighteness, and tone curve transformations.}
\label{f_distortion}
\end{figure*}

\begin{table}[t]
\centering
\caption{Dataset statistics. The validation stage only involves simple distortions. The test stage further includes complex distortions. The distortion in Validation-2 is harder than in Validation-1.}
\begin{tabular}{c|c|c|c}
\toprule[1.2pt]
\textbf{Dataset} & \textbf{Images}  & \textbf{Generators} & \textbf{Distortions} \\
\hline
Toy     & 10,000    & 10 & 0 \\
Train           & $~$277,000  & 20 & 0 \\
Validation-1  & 10,000    & 25 & 5 \\
Validation-2 & 2,500     & 25 & 5  \\
Public test     & 2,500     & 30 & 7 \\
Private test    & 2,500     & 35 & 9  \\
\bottomrule[1.2pt]
\end{tabular}
\label{tab:dataset_splits}
\end{table}

\subsection{Parameter-Efficient Fine-Tuning}

Adapting massive foundation models to many downstream tasks presents a significant computational bottleneck. Traditional full fine-tuning strategy needs to update billions of parameters, which is not only computationally prohibitive but also frequently leads to catastrophic forgetting \cite{houlsby2019adapters,lora}. To circumvent this, Parameter-Efficient Fine-Tuning (PEFT) methods have been widely adopted. Low-Rank Adaptation (LoRA) is arguably the most prominent of these techniques \cite{lora}. LoRA hypothesizes that the updates to the weight matrices during fine-tuning have a low intrinsic rank \cite{houlsby2019adapters,lora,zhang2023adalora}. Therefore, it freezes the pre-trained parameters and incorporates trainable rank decomposition matrices into specific layers-typically the attention mechanisms. Therefore, applying LoRA allows us to surgically adapt the visual foundation model to the AIGI detection task with minimal computational overhead, thereby preserving its detection performance against unseen data \cite{lora,jia2022visual}.

\subsection{Robust detection in the wild.}

Recently, AIGI detection has shifted from pursuing high accuracy in ideal settings to ensuring robustness against real-world perturbations. Early studies primarily relied on data augmentations (e.g., blurring, JPEG compression) and ensemble learning to enhance generalization against unseen generators and standard post-processing \cite{wang2020cnn,mandelli2022detecting}. However, the rapid emergence of diffusion models revealed that legacy GAN-based detectors suffer severe performance degradation under complex social media transmission scenarios \cite{corvi2023detection}. Consequently, recent efforts address this challenge through two main avenues: leveraging robust pre-trained visual representations to improve cross-paradigm out-of-distribution (OOD) generalization \cite{gaintseva2024improving}, and constructing rigorous in-the-wild benchmarks to empirically evaluate detector vulnerabilities in open-world environments \cite{konstantinidou2025navigating}.
\section{Data Analysis}
\label{sec:analysis}

Prior to introducing the proposed approach, we first conduct a comprehensive analysis of the dataset provided in this competition. Tab.~\ref{tab:dataset_splits} summarizes the data distribution across the toy, training, validation, and test sets. Notably, as the competition progresses, the difficulty of the detection task increases substantially, driven by the growing diversity of generative models and the increasing complexity of applied distortions. As shown in Fig.~\ref{f_distortion}, different from previous AIGI detection, which only distorted the images with a single transformation with marginal strength, the images in this competition are distorted with the combination of multiple distortions and high-level strength, which incurs great challenges to previous detection methods.

\begin{figure}[t]
\centering
\includegraphics[width=\linewidth]{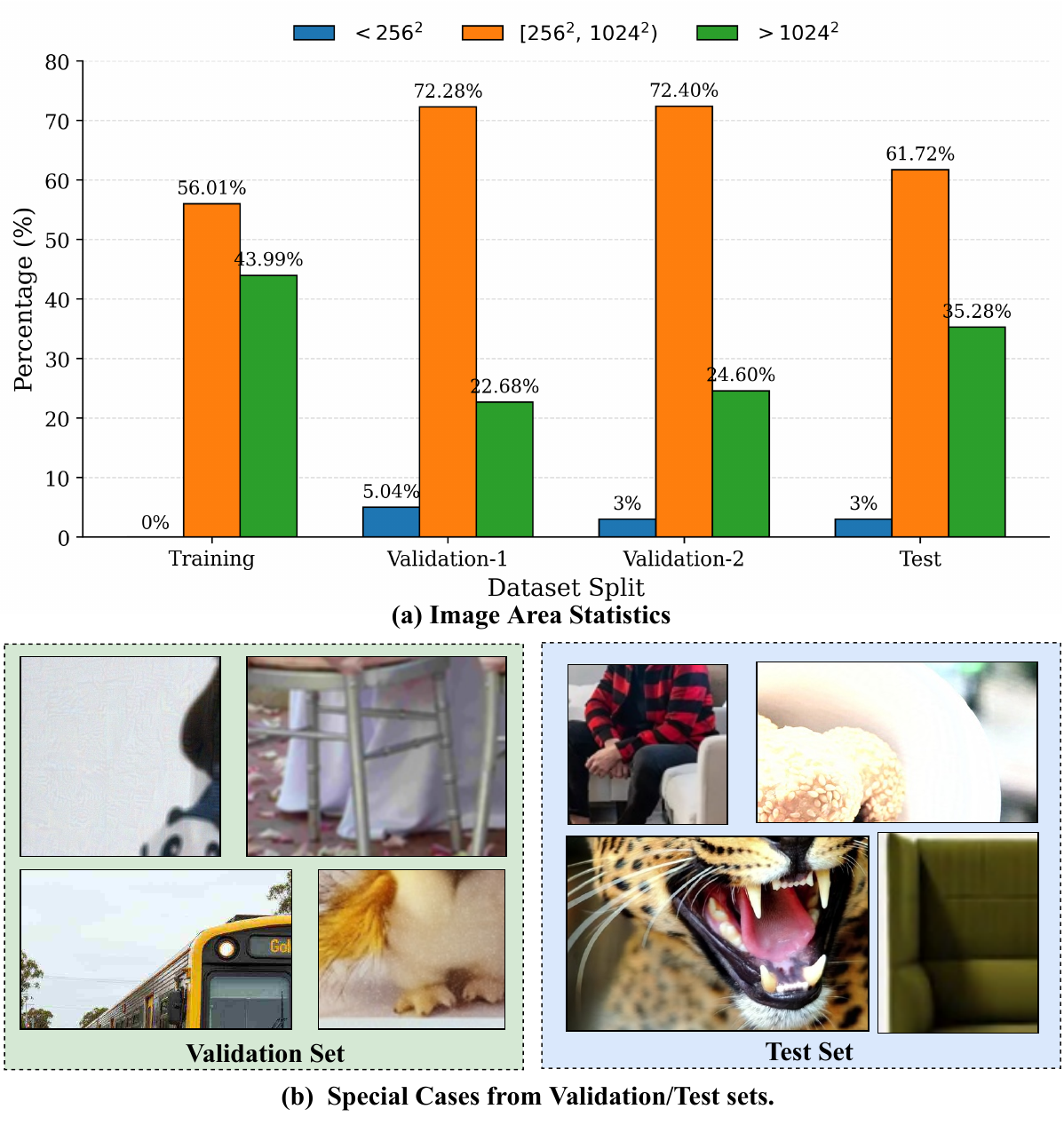}
\caption{\textbf{(a) Image Area Statistics.} There is a distribution mismatching between the training and validation/test sets. \textbf{(b) Special Cases from Validation/Test sets.} Some images only appear to have local semantic or non-semantic information.}
\label{f_area_distribution}
\end{figure}

According to the data provided by the organizer, there are no distortions related to the training sets, which means the distortion simulation is important to achieve the robust detection performance during the training stage. Besides, Fig.~\ref{f_area_distribution}(a) depicts statistics related to image area from each set. It can be observed that there is a distribution mismatch between the training and validation/test sets. Specifically, the images in the training set often exhibit complete semantic information, but the validation/test sets appear to have local semantic or non-semantic information (see Fig.~\ref{f_area_distribution}(b)). Furthermore, the ratio between the real and fake images exhibits imbalance, which leads to bias during detection.

Considering the observed characteristics, we suggest that robust AIGI detection can be achieved through the data distribution simulation. Such a simulation includes the various distortions and sizes, thereby reducing the discrepancy between training and test domains. Besides, given the strong semantic representation capabilities of visual foundation models, fine-tuning these models provides a promising way for AIGI detection. Finally, to preserve accuracy on clean samples while maintaining resilience to distorted inputs, the training process should simultaneously consider the detection generalization and robustness.
\section{Methodology}
\label{sec:methodology}

\subsection{Overall Framework} 
\begin{figure*}
\centering
\includegraphics[width=\linewidth]{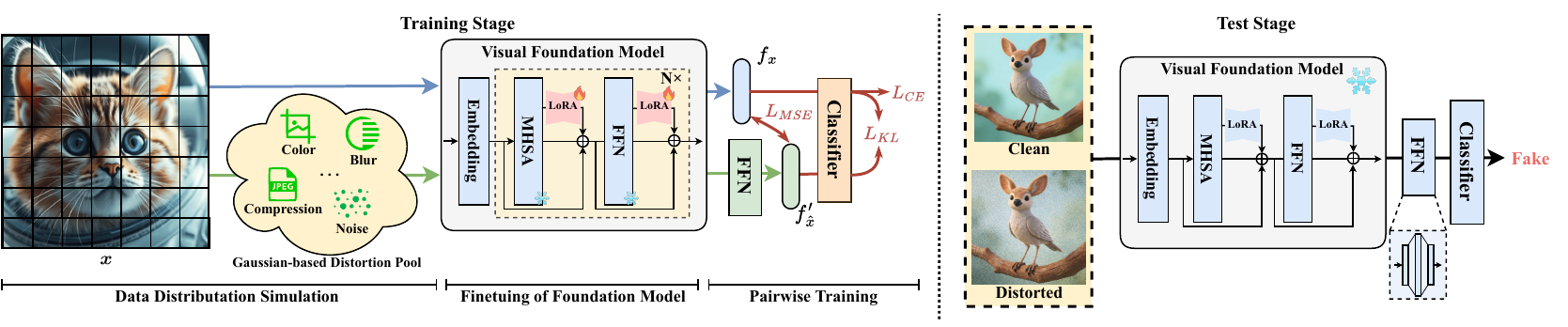}
\caption{\textbf{The overall framework of LPT.} To achieve robust detection, our strategy involves the simulation of data distribution, the fine-tuning of the visual foundation model, and the pairwise training, respectively.}
\label{f_arc}
\end{figure*}

The framework of LPT is illustrated in Fig.~\ref{f_arc}. During the training stage, the images are augmented to simulate the data distributions from the validation and test sets. To better perceive the trace from AIGI, we utilize the visual foundation model and finetune it to adapt the detection task. Finally, the pairwise training strategy is adopted to simultaneously train the detector with the images under the clean and distorted situations, which makes the detector adapt to the different generation models and distorted situations.

\subsection{Data Distribution Simulation} 
The data simulation involves two key components, i.e., image distortions and sizes, which are essential for achieving robust and generalized detection performance. 

The first critical step in LPT is to simulate the adverse conditions in the wild. Initially, we apply the standard baseline distortions provided by the competition organizers, i.e., each image is distorted with random combinational transformations, and the strength of each transformation has five levels. The level selection follows Gaussian sampling as:
\begin{equation}
P(x)=\frac{1}{\sigma \sqrt{2 \pi}} e^{-\frac{(x-\mu)^2}{2 \sigma^2}},
\end{equation}
where $x$ denotes the index of the strength levels of distortion. $\mu$ and $\sigma$ indicate mean value and standard deviation. However, empirical observation of the validation set reveals a significant distribution shift. To rectify this, we further introduce several complex types of distortions, such as speckle noise, tone curve, organic moire, and color jitter. These specific distortions can better fit the expected data distribution of real-world social media platforms. Furthermore, to forcefully decouple the model's reliance on fragile, easily-corrupted features, we artificially increase the strength levels of the distortions by setting $\mu=3$. The introduced distortions are listed in Tab.~\ref{tab:distortions}. 

\begin{table}[t]
\centering
\caption{The introduction of distortions during the model training.}
\label{tab:distortions}
\setlength{\tabcolsep}{0.35cm}{
\begin{tabular}{lc}
\toprule
% \textbf{Participants} & \textbf{Public Test Hard Set} & \textbf{Private Test Hard Set} \\
\textbf{Type}  & \textbf{Distortions} \\
\midrule
Blur & Gaussian~/~Lens~\\
Noise & Gaussian~/~Impulse~/~Speckle\\
Color & Shift/~ColorJitter~/~Moire/~Tone\\
Brightness  & Brighten~/~Darken \\
Compression & JPEG~/~Quantization\\
Spatial Space & Jitter \\ 
\bottomrule
\end{tabular}}
\end{table}

The next step is to simulate the various sizes from the validation set. Algorithm~\ref{alg:random_crop_resize} describes the combination of random crop and resize augmentation. Considering the images from the training set are always represented with a large size, we thus crop or resize the images to fit the data from the validation set. Specifically, after obtaining a training image, we randomly crop the image into two different sizes. These sizes can be divided into two categories, i.e., larger and smaller than the target sizes, which fit the images with local semantic and non-semantic information. Furthermore, we observe that some validation images are presented with a resize format; the training images are also randomly resized with different interpolation modes.

\begin{algorithm}[t]
\caption{Random Crop and Resize Augmentation}
\label{alg:random_crop_resize}
\begin{algorithmic}[1]
\REQUIRE Input image $\mathbf{img}$ of shape $(H,W,C)$, target size $\texttt{tgt}$, thresholds $T_1$, $T_2$
\ENSURE Augmented image $\texttt{img}$

\STATE $(H,W,C) \gets \texttt{img}.\texttt{shape}$
\STATE $r_c \gets$ random sampling from $[0,1]$
\STATE $r_r \gets$ random sampling from $[0,1]$
\STATE $\mathcal{I}\gets\{\texttt{NEAREST},\texttt{LINEAR},\texttt{CUBIC},\texttt{AREA}\}$

\IF{$r_c < T_1$}
    \STATE $r \gets $ random sampling from [0,1]
    \STATE $H_{\text{new}} \gets
    \begin{cases}
    \texttt{randint}(64,\texttt{tgt}),  H\ge \texttt{tgt},r<T_2\\
    \texttt{randint}(\texttt{tgt},H),  H\ge \texttt{tgt}\\
    H,  \text{otherwise}
    \end{cases}$

    \STATE $W_{\text{new}} \gets
    \begin{cases}
    \texttt{randint}(32,\texttt{tgt}),  W\ge \texttt{tgt},r<T_2\\
    \texttt{randint}(\texttt{tgt},W), W\ge \texttt{tgt}\\
    W, \text{otherwise}
    \end{cases}$

    \STATE $\texttt{top} \gets \texttt{randint}(0,\max(H_{\text{new}}-\texttt{tgt},0))$
    \STATE $\texttt{left} \gets \texttt{randint}(0,\max(W_{\text{new}}-\texttt{tgt},0))$
    \STATE $\texttt{crop\_h} \gets \min(H_{\text{new}},\texttt{tgt})$
    \STATE $\texttt{crop\_w} \gets \min(W_{\text{new}},\texttt{tgt})$
    \STATE $\texttt{img} \gets \texttt{img}[\texttt{top}:\texttt{top}+\texttt{crop\_h},\ \texttt{left}:\texttt{left}+\texttt{crop\_w}]$

\ELSE
    \IF{$r_r < 0.25$}
        \STATE $\texttt{mode} \gets \texttt{randchoice}(\mathcal{I})$
        \STATE $\texttt{img} \gets \texttt{resize}(\texttt{img},(W/2,H/2),\texttt{mode})$
    \ENDIF
\ENDIF
\end{algorithmic}
\end{algorithm}

\subsection{Model and Fine-tuning Component Selection}

The backbone foundation model utilized in the LPT framework is EVA-CLIP \cite{eva}. Given the strong semantic aware capability inherently present in large-scale visual foundation models, it is highly effective to finetune these models to perceive the subtle traces left by AIGI. While EVA-CLIP possesses strong visual representations learned during the massive pre-training phase, it must be meticulously adapted to the specific nuances of generated imagery. A critical challenge in this adaptation process is avoiding catastrophic forgetting—a phenomenon where the model loses its pre-trained generalized knowledge and overfits narrowly to the new training data. To achieve the AIGI detection without sacrificing the model's foundational ability, we implement a parameter-efficient fine-tuning strategy utilizing LoRA \cite{lora}. Instead of updating the billions of parameters within the entire architecture, LoRA freezes the pre-trained weights and injects trainable rank decomposition matrices into specific layers. Specifically, the fine-tuning components within our adapted EVA-CLIP consist of the linear layers located within the multi-head self-attention (MHSA) module, as well as the feed-forward network (FFN) present at each visual transformer block. This fine-tuning strategy allows the network to adapt to the AIGI detection task with minimal computational overhead, ensuring that it preserves its inherent representation against unseen data distributions.

\subsection{Pairwise Sample Training}
A common phenomenon in robust model training is that heavy data augmentation, while necessary for in-the-wild detection, often causes degradation in performance when evaluating clean samples \cite{trinh2024improving,saikia2021improving}. To ensure that the detection performance remains optimal under clean conditions, the training process must simultaneously consider both detection generalization and environmental robustness. To resolve this, we adopt a novel pairwise training strategy. This strategy is achieved by introducing both the clean samples and their corresponding distorted counterparts simultaneously within each training batch. During the forward pass, the foundation model extracts representations for both inputs. However, because distortions shift the feature distribution, the features extracted from the distorted images are dynamically passed through and corrected by a dedicated auxiliary FFN. Similar to the FFN located in the visual foundation model, there are several Fully Connected (FC) layers and the structure presents invert bottleneck type. The the hidden dimension in the intermediate layer set to twice that of the first and third layers. This auxiliary network acts as a semantic anchor, explicitly forcing the corrupted features to map closer to their pristine counterparts. Consequently, to optimize both the detection accuracy and the feature-level alignment, the entire network is trained using a comprehensive joint loss function $\mathcal{L}$ as follows:

\begin{equation}
\mathcal{L}=\mathcal{L}_{C E}\left(\bm{x},\bm{y}\right)+\alpha\cdot\mathcal{L}_{KL}\left(\bm{x},\bm{\hat{x}}\right)+\beta\cdot\mathcal{L}_{MSE}\left(\bm{f}_{x},\bm{f}'_{\hat{x}}\right),
\end{equation}
where $\bm{x}$ and $\bm{\hat{x}}$ indicate the clean samples and the distorted samples, respectively. The terms $\bm{f}_{x}$ and $\bm{f}_{\hat{x}}^{\prime}$ represent the high-dimensional clean sample features and the distorted sample features after correction, respectively. Furthermore, $\mathcal{L}_{CE}$, $\mathcal{L}_{KL}$, and $\mathcal{L}_{MSE}$ denote the cross-entropy loss for accurate detection, the Kullback-Leibler divergence for aligning the predictive output distributions, and the mean square error for minimizing the distance between the feature representations. To optimally balance these training objectives, $\alpha$ and $\beta$ are explicitly set to 0.5 and 0.25, respectively.
\section{Experiments}
\label{sec:experiments}

\subsection{Experimental Setup}
\textbf{Training Details.} The training epoch equals 5. The learning rate is initialized as 2e-4 with the AdamW\cite{adamw} optimizer and cosine annealing scheduler\cite{cos}. The weight decay is set as 5e-4. Moreover, we additionally introduce So-Fake\cite{sofake} and Chameleon\cite{chameleon} datasets during the final training stage to further improve the generalization of the detector. $T_1$ and $T_2$ are both set to 0.3. All experiments are conducted on Pytorch 2.7.1 and 8 NVIDIA A800 GPUs. In the LoRA module, both the scaling factor and the rank are set to 16. All images are trained with 224$\times$224, and each test image is resized to the same resolution during inference. 

\noindent\textbf{Evaluation Metric.} Accuracy (Acc) and Area Under the Curve (AUC) have been employed when comparing our approach with the state-of-the-art (SOTA) methods and evaluating the effectiveness of each proposed component. The AUC measures the ability of a detector to discriminate between fake and real classes. The AUC is
the area under the Receiver Operating Characteristic (ROC)
curve, which shows the TP (True Positive) rate against the FP (False Positive) rate at various thresholds.

\begin{table}[t]
\centering
\caption{Performance comparison on the challenge test set (AUC \%). Our approach secures the third placement. `\textbf{Hard}' denotes that the results are computed from the distorted images only.}
\label{tab:challenge_test}
\setlength{\tabcolsep}{0.2cm}{
\begin{tabular}{lcc}
\toprule
% \textbf{Participants} & \textbf{Public Test Hard Set} & \textbf{Private Test Hard Set} \\
\multirow{2}{*}{\textbf{Participants}}  & \textbf{Public} & \textbf{Private} \\
& \textbf{Test Hard Set} & \textbf{ Test Hard Set}\\
\midrule
MICV & 97.38 & 97.22\\
Ant International & 97.30 & 97.20 \\
INTSIG           & 90.78  & 91.30\\
vincentlc  & 86.42  & 87.33 \\
Reagvis Labs & 85.66 & 86.00 \\ 
\textbf{Ours} & \textbf{92.15} & \textbf{92.50} \\
\bottomrule
\end{tabular}}
\end{table}

\subsection{Comparison with Competition Teams}
We compare our approach against many competitors on the challenge test set. The results are summarized in Tab.~\ref{tab:challenge_test}.
The LPT approach demonstrated exceptional robust detection capabilities, yielding an Area Under the Curve (AUC) of 92.15\% and 92.50\% on the public and private test hard set, respectively. Notably, our approach consistently outperformed the majority of contemporary ensembles, securing a top-three placement in the challenge. The strong performance, particularly in AUC, highlights the model’s ability to make reliable detection under the distorted scenarios.

\subsection{Ablation Study}
In the absence of specific instructions, we use the pretrained CLIP L-14-336 \cite{clip} as the visual foundation model to evaluate the effectiveness of each proposed component. 

\subsubsection{Pairwise Sample Training}
% 成对样本 vs 一般训练

\begin{table}[t]
\centering
\caption{Ablation study on training strategy (AUC \%). Each $\checkmark$ indicates that the component is included. `*' means the results are the average from Validation-1 and Validation-2 sets. All results are computed from the clean and distorted images.}
\label{tab:ablation_strategy}
\begin{tabular}{ccccc}
\toprule
$\mathcal{L}_{CE}$ & $\mathcal{L}_{KL}$ & $\mathcal{L}_{MSE}$ & \textbf{Validation*} & \textbf{Public Test} \\
\midrule
$\checkmark$ & - & - & 97.31 & 87.91 \\
$\checkmark$ & $\checkmark$ & - & 97.38 & 88.06 \\
$\checkmark$ & $\checkmark$ & $\checkmark$ & \textbf{97.79} & \textbf{89.26} \\
\bottomrule
\end{tabular}
\end{table}

We first analyze the effectiveness of the proposed training strategy. Tab.~\ref{tab:ablation_strategy} lists different configurations related to the model training. When trained solely with the $\mathcal{L}_{CE}$, both clean and distorted images are treated under a unified optimization scheme. However, since the objective is to develop a detector that is both robust and generalizable, this formulation leads to an inherent entanglement between generalization and robustness objectives. As the AI-generated traces are subtle, some image distortions will interfere with these traces and increase the detection difficulties. This training strategy makes the model focus on the image distortions, traces, or the distorted AI-generated traces, thereby impacting the model convergence. Consequently, when presented with clean images or unseen distortion patterns, the detection performance degrades. By introducing a decoupled optimization scheme via KL divergence, the detector can clearly capture the AI-generated traces from the clean images and the varied versions after image distortions. The results from the second row show that the detection performance is improved. Furthermore, by incorporating additional alignment in the high-dimensional feature space, the model achieves the best detection performance, highlighting the importance of decoupling and feature consistency.

\subsubsection{Different Sample Simulation}
\begin{figure}[t]
\centering
\includegraphics[width=\linewidth]{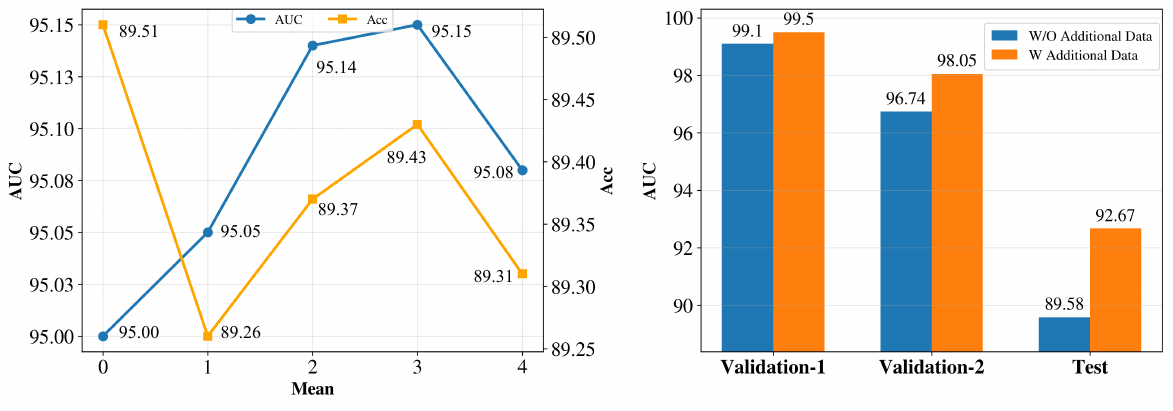}
\caption{\textbf{Left}: The impact of the distortion strength. \textbf{Right}: The impact of additional data incorporation. }
\label{f_stength}
\end{figure}

% 数据模拟（裁剪/resize） 失真模拟（额外失真/失真强度）
To evaluate the effectiveness of the data distribution simulation, we conduct ablation studies on both data scale variation and image distortion modeling. Tab.~\ref{tab:ablation_simulation} shows the ablation studies related to the image size and distortion. Considering the size variations in the validation and test sets, we further simulate the data size with Algorithm~\ref{alg:random_crop_resize}. The results in the second column show the effectiveness of the proposed size simulation. Given the absence of image distortions in the training set, we further incorporate various image distortions based on the supplementary codes and increase the distortion strength. The results in the final column show the effectiveness of the fitted distortions. Besides, we also evaluate the detection performance under varying distortion strengths. The variation from the left panel of Fig.~\ref{f_stength} shows that the initially increasing strength can improve the detection performance. This is because moderate distortion simulation better aligns the training distribution with that of the validation and test sets. However, when further increasing the strength, the performance appears to decrease due to the mismatch of practical distortions and the increased difficulty of learning from severely corrupted samples.

\begin{table}[t]
\centering
\caption{Ablation study on the data distribution simulation (AUC \%). All results are computed from the clean and distorted images.}
\label{tab:ablation_simulation}
\setlength{\tabcolsep}{0.15cm}{
\begin{tabular}{cccccc}
\toprule
\textbf{Size} & \textbf{Distortion} & \textbf{Val.-1} & \textbf{Val.-2} & \textbf{Public Test} & \textbf{Avg.} \\
\midrule
- & - & 98.89 & 96.69 & 89.26 &94.95 \\
$\checkmark$ & - &  \textbf{99.17} & \textbf{96.77} & 89.27&95.07 \\
$\checkmark$ & $\checkmark$ &  99.10 & 96.74 & \textbf{89.58}&\textbf{95.15} \\
\bottomrule
\end{tabular}}
\end{table}

\subsubsection{Data Extension}
The right of Fig.~\ref{f_stength} shows the benefits of the additional data extension. Considering the limited generation methods in the Validation-1 set, detection performance on this split is close to saturation, and the remaining errors are largely attributable to image distortions rather than insufficient coverage of generative traces. However, the test set encompasses a substantially broader range of generation methods than either validation split. Incorporating other datasets during training can improve generalization and leads to a pronounced performance gain on the test set.

\subsubsection{Foundation Model Selection}

\begin{table}[t]
\centering
\caption{Ablation study on the visual foundation model selection (AUC \%). Meta-CLIP2 \cite{meta2} is the Big-G version, and EVA-CLIP \cite{eva} is the Big-E version. All results are computed from the clean and distorted images. `*' denotes the utilization of data extension.}
\label{tab:ablation_model}
\setlength{\tabcolsep}{0.09cm}{
\begin{tabular}{cccc}
\toprule
\textbf{Model}  & \textbf{Validation-1} & \textbf{Validation-2} & \textbf{Public Test} \\
\midrule
CLIP* & 99.50  & 98.05& 92.67 \\
Meta-CLIP2* & \textbf{99.60}  & \textbf{98.44} & \textbf{95.73} \\
EVA-CLIP* &  99.22 & 98.06 & 95.15 \\
\bottomrule
\end{tabular}}
\end{table}

To evaluate the improvements of the selection from different visual foundation models, Tab.~\ref{tab:ablation_model} shows the ablation studies with three models, i.e., CLIP \cite{clip}, Meta-CLIP2 \cite{meta2}, and EVA-CLIP \cite{eva}. The results show that Meta-CLIP2 and EVA-CLIP can both get significant improvements compared with the CLIP framework. As to the Meta-CLIP2, it focuses on rigorous data engineering and global applicability. EVA-CLIP \cite{eva} prioritizes extreme model scaling and training efficiency. By leveraging progressive weak-to-strong initialization, scaling parameters up to 18 billion, and employing advanced optimization techniques such as token dropping. Therefore, the scale of model parameters, the surge in pre-training data volume, and the optimization in the training process bring benefits to the visual representation and thereby improve the AIGI detection. Considering that the robust performance in EVA-CLIP is better than Meta-CLIP2 (92.15\% vs. 91.89\% under private hard test set), we use EVA-CLIP as the default foundation model.

\begin{table}[t]
\centering
\caption{Performance comparison (Acc \%) on the ForenSynths\cite{wang2020cnn}. Except the `\textbf{Avg.}' column, each result is the average detection accuracy from the eight generation methods (`ProGAN', `CycleGAN', `Deepfake', `GauGAN', `BigGAN', `StarGAN', `StyleGAN', and `StyleGAN2'). Here, Q, k, $\sigma$, and v denote quality factor, kernel size, stand derivation, and variation.}
\label{tab:sota_compare}
\setlength{\tabcolsep}{0.15cm}{
\begin{tabular}{lccccc}
\toprule
% \textbf{Participants} & \textbf{Public Test Hard Set} & \textbf{Private Test Hard Set} \\
\multirow{2}{*}{\textbf{Methods}}  & \multirow{2}{*}{\textbf{Clean}} & \textbf{JPEG} & \textbf{Noise} & \textbf{Blur}& \multirow{2}{*}{\textbf{Avg.}}\\
&&Q=50&k=7 $\sigma$=2&v=25&\\
\midrule
NPR~\cite{tan2024rethinking} &91.80	&50.20	&50.00	&65.70	&64.42 \\
MLEP~\cite{yuanmlep} &89.85	&48.90	&49.90	&74.90	&65.88 \\
CNN~\cite{wang2020cnn} &74.91	&57.57	&70.48	&74.80	&69.44\\
Effort~\cite{effort} & \textbf{97.22}&	60.32&	68.56&	90.88 & 79.24\\
DDA~\cite{chen2025dual}	&84.16	&78.02	&79.65	&83.40	&81.30 \\
RINE~\cite{koutlis2024leveraging} &95.40 & 81.50 & 84.60 & 91.70 & 88.30\\
\textbf{Ours} & 96.01	&\textbf{90.38}	&\textbf{93.67}	&\textbf{95.23}	&\textbf{93.82}\\
\bottomrule
\end{tabular}}
\end{table}

\subsection{Comparison with SOTA methods}
\par To evaluate the detection performance under the standard protocol, we follow previous works\cite{tan2024rethinking,yuanmlep,koutlis2024leveraging} by adopting specific 4-class samples from ProGAN during training. Tab.~\ref{tab:sota_compare} reports a comparison against SOTA methods on ForenSynths\cite{wang2020cnn}. Although several previous methods attain the promising performance under the clean condition, their performance degrades noticeably under distortions. In contrast, LPT achieves advanced generalized and robust performance under the clean and distorted cases, highlighting its superior generalization and robustness and underscoring its practical value for open-world deployment.
\section{Conclusion and Future Work}
\label{sec:conclusion}

\textbf{Conclusion.} We introduced the LPT strategy to combat the severe performance degradation experienced by AI-Generated Image detectors in wild environments. By combining targeted fine-tuning of the EVA-CLIP foundation model with a rigorous pairwise training constraint and customized data distortion simulation, our approach effectively corrects corrupted feature representations. Achieving a top-three placement in the NTIRE 2026 challenge demonstrates the generalization and robustness of LPT. 

\noindent\textbf{Future Work.}  To improve the robustness, our future work mainly involves three aspects. First, to address variations in image size, we will integrate multi-scale features across different Transformer blocks to optimize results. Second, we will bridge the gap between training and real-world degradation via subjecting the detector and distortions to joint adversarial training. The distortion set learns to simulate perturbations to which the detector is most vulnerable, thereby systematically bolstering robustness. Third, considering that different distortions impact the AI-generated traces with varying degrees, we will employ a Mixture-of-Experts (MoE) architecture to better adapt to diverse distortions.
{
    \small
    \bibliographystyle{ieeenat_fullname}
    \bibliography{main}
}


\end{document}